# ControlMap: Controllable High-Definition Map Generation for Traffic Scenario Simulation

Marwan Farag[1], Steffen Wäldele[2], and Yu Yao[3]

[1] University of Stuttgart, Stuttgart, Germany
marwan.farag@iss.uni-stuttgart.de
[2] Robert Bosch GmbH, Renningen, Germany
steffen.Waeldele@de.bosch.com
[3] Motional, Inc
yu.yao@motional.com

**Abstract.** Simulation is central to validating autonomous driving systems, yet current pipelines are limited by insufficient scenario diversity due to costly High Definition (HD) map creation. Scaling HD maps requires expensive data collection and manual processing. Moreover, existing generative models lack the fine-grained control necessary to target specific road topologies during generation.
This paper presents a data-driven pipeline for controllable HD map generation using latent diffusion and ControlNet for spatial conditioning. To our knowledge, we are the first to inject spatial guidance signals into a diffusion model for HD map synthesis. Furthermore, our model supports adjustable conditioning strength through classifier-free guidance and city-level style transfer via city label conditioning. To complement existing metrics, we introduce two novel metrics to evaluate adherence to the control signal and similarity to ground-truth maps. Experiments demonstrate that our model generates realistic HD maps that faithfully follow input road topologies while accurately preserving city-specific details.

## 1 Introduction

Simulation pipelines are a cornerstone of the verification process for the safety-critical components of autonomous driving systems. However, their utility relies on access to realistic and diverse driving scenarios [13, 26]. At the core of each driving scenario lies a *High Definition (HD) map* that includes detailed lane geometry, traffic infrastructure, and intersection layout [2,13]. Creating these maps requires specialized sensor-equipped vehicles, extensive validation, and manual annotation [8, 26]. This makes the process very costly and difficult to scale [14]. Large benchmarks such as nuPlan [3], which collected over 1300 hours of driving logs and more than 2TB of sensor data, still cover only a limited set of regions.

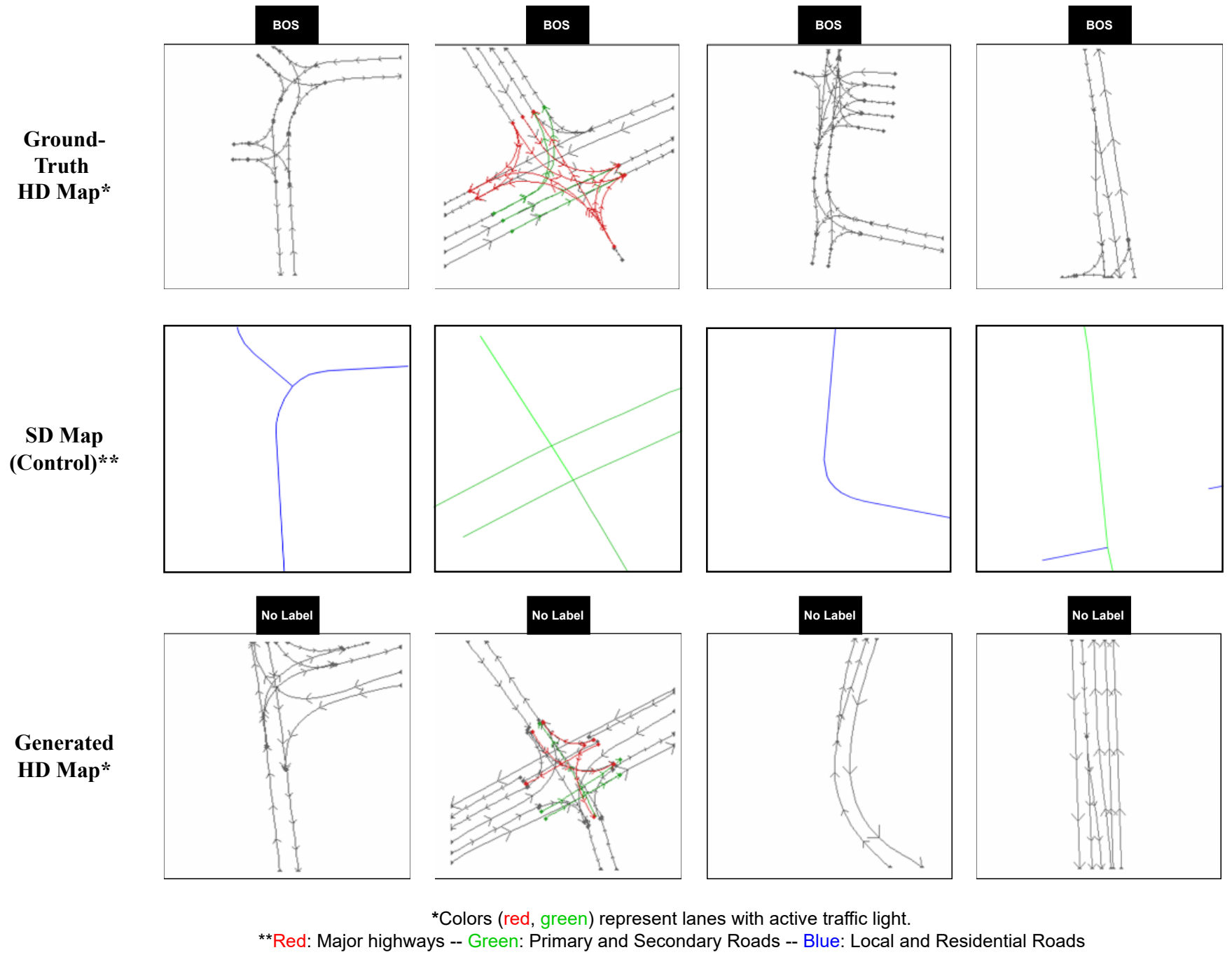

**Fig. 1:** We condition on SD maps that are cheap to obtain (*e.g.*, from OpenStreetMap [19]) to generate diverse HD maps with user-specified structure. Our generations (bottom) follow the control layout (middle) and recover realistic HD detail consistent with test ground truth (top). *BOS* denotes ground-truth maps coming from Boston; *No Label* indicates that the model took no city label as input during generation.

This underlines the challenge of extending HD map-based simulation to new environments [26].

Another challenge is the dependence on a limited diversity of scenarios for testing. Autonomous systems must be tested across an extremely wide range of situations, including rare events and complex interactions that might not appear in existing datasets [3,7]. Most simulation pipelines depend on static maps or pre-recorded logs, which inherently limit scenario variability [2]. Data-driven map generation offers a promising alternative. Recent generative methods train diffusion models [10] to synthesise novel and diverse driving scenes that circumvent the real-world data limitations [5,16,21]. It is expected that such methods will widen the scope of simulation-based testing for autonomous systems [5]. However, current approaches to autonomous driving map generation generally lack user-specified layout control. Consequently, there is a strong motivation for methods that can provide precise structural guidance, particularly for rare and challenging scenarios, without the expense and rigidity of HD maps.

We propose using Standard Definition (SD) maps (e.g., OpenStreetMap [19]) to guide generation. SD maps offer a lightweight topological skeleton of lanes and intersections, providing sufficient structural guidance for generation while omitting costly HD-level details. Therefore, we can summarize our contributions as follows:

- We introduce *ControlMap*, which, to the best of our knowledge, is among the first frameworks for spatially conditioned HD map generation. Our approach transforms publicly available SD map data [19] into lightweight yet structured rasterized control signals, enabling high-fidelity HD map synthesis. This mitigates the scalability challenges of costly HD map acquisition while enabling controllable generation of diverse map layouts.
- To address the lack of controllability metrics in prior work, we introduce novel metrics evaluating semantic and topological consistency between generated maps, SD inputs, and ground truth. These demonstrate our conditioned approach aligns closer to ground truth than unconditional baselines.
- We use classifier-free guidance [11] to serve as a continuous control mechanism for conditioned HD map generation, providing a flexible knob to balance adherence to the SD control signal with generative diversity.

## 2 Related Work

### 2.1 Scenario Generation

Conventional simulators such as CARLA [7], nuPlan [3], and Waymax [9] rely on prerecorded scenes to initialize simulations. This reliance limits scalability and reduces coverage of edge-case traffic scenarios. To address this issue, data-driven scene generation trains models on real map and sensor data [1, 3], enabling the synthesis of diverse and realistic scenarios. Diffusion-based approaches [5] and vectorized scene generators [21] are promising directions for producing structured traffic scenes at scale.
We propose a taxonomy that categorizes scenario generation methods by their inputs and outputs.
**Generating Only HD Maps:** HDMapGen [17] generates hierarchical autoregressive lane graphs that capture global connectivity and local geometry.
**Generating Agents Conditioned on an HD Map:** SceneGen [24] places agents sequentially using an autoregressive model conditioned on an HD map and ego context, while SceneControl [16] uses latent diffusion with semantic tokens to guide the generation of agent positions and trajectories.
**Jointly Generating Agents and HD Maps:** SLEDGE [5], DriveSceneGen [23], and Scenario Dreamer [21] synthesize full traffic scenes by generating HD maps and placing traffic agents onto those maps. However, these methods do not provide explicit control over the spatial layout of the generated HD map.
Our work represents a fourth category: **HD Map Generation Conditioned on an SD Map Control Signal**. To the best of our knowledge, we are the first to explore this direction for diffusion-based HD map generation. In addition, because our method builds on the SLEDGE framework [5], extending it to joint agent and HD map generation is straightforward.

### 2.2 Conditioning in Diffusion Models

Conditioning provides a mechanism to guide diffusion models toward desired outputs. Prior work uses cross-attention, as in DALL·E [20] and Imagen [22], lightweight adapters [18], and classifier-free guidance [11]. However, these techniques provide limited spatial precision. ControlNet [27] addresses this limitation by adding zero-initialized convolutional branches to a U-Net–based diffusion backbone, encoding structured control signals that enable fine-grained geometric and semantic alignment while preserving the pretrained model's generalization.

In addition, recent work has shown that ControlNet-style conditioning can be extended beyond U-Net backbones to Diffusion Transformers (DiT) by adding a dedicated control branch whose outputs are injected into the frozen DiT blocks through *zero-initialized linear layers*, preserving the pretrained behavior at initialization. This has proven effective for controllable music generation [12] (*e.g.*, melody- and text-guided control) and text-to-image synthesis [4].

Inspired by these works, we integrate ControlNet-style conditioning into the latent-space diffusion transformer of SLEDGE, to enable SD map-based spatial conditioning for HD map generation. In our setup, the SD map control signal guides the global layout of the generated HD map, while the diffusion model fills in the lane-level detail.

## 3 Method

### 3.1 SLEDGE Baseline

We adopt SLEDGE [5], a diffusion-based generative model for traffic scenes and HD maps as our baseline. SLEDGE represents each driving scene as a symbolic *SLEDGE Vector*, which is rasterized into a multi-channel *Rasterized State Image (RSI)* and encoded by a ResNet-based RVAE into a compact latent map. A Diffusion Transformer (DiT) is then trained to generate these latent representations, which are decoded back into structured HD map and scene elements by the frozen RVAE decoder. SLEDGE evaluates different DiT backbone sizes, reporting $\sim$138M parameters for DiT-B and $\sim$487M parameters for DiT-XL. For a more detailed discussion of the original SLEDGE architecture, the interested reader is referred to [5].

In SLEDGE, the HD map generator is only weakly conditioned: the model receives a city label to disambiguate city-specific topology, but does not incorporate any spatial conditioning signal. Specifically, the city label is provided as a one-hot vector with dimension equal to one plus the number of cities in the training set. The additional label is used during training for the purpose of city label drop-out and during inference for generating an output which is not associated with any specific city from the training set.

### 3.2 ControlMap: Introducing SD Map Conditioning

Just as the DiT-part of the SLEDGE model, our SD map conditioned version operates in the latent space of the RVAE and is trained using the Denoising Diffusion Probabilistic Model (DDPM) objective [10]. An overview of the architecture of our extended ControlMap model based on the DiT-B ($\sim$138M parameters) version is illustrated in Fig. 2. Each SD map is rasterized as a $128\times128$ RGB image and processed through a convolutional control encoder composed of 4 residual Conv$\rightarrow$GroupNorm$\rightarrow$GELU blocks with strided downsampling. The output is a $64\times8\times8$ tensor augmented with 2D sinusoidal positional embeddings to preserve spatial locality when flattened into tokens for transformer input. Details on SD map image construction are provided in the supplementary material.

Conditioning is applied to the first $M{=}6$ transformer blocks by attaching parallel trainable ControlNet copies of these blocks, initialized from the pretrained weights. The outputs of the ControlNet [27] blocks are processed through a zero-initialized linear projection and additively injected into the frozen DiT blocks, forming a residual modulation pathway. The remaining blocks stay frozen to preserve the pretrained generative prior. After the final block, the output tokens are projected and unpatchified into an $64\times8\times8$ latent tensor, decoded by the pretrained RVAE. Following SLEDGE, city-label embeddings are further injected via adaptive normalization [25], enabling conditioning on urban style (*e.g.*, Las Vegas vs. Singapore).

### 3.3 Metrics for Control Fidelity Evaluation

Since previous work did not focus on spatially conditioned HD map generation, existing metrics such as Fréchet Distance [6,17] focus on comparing distribution-level properties. While useful, these metrics are ineffective when evaluating the quality of the spatial conditioning on a sample-level. Therefore, we introduce two additional metrics called *Control Recall* and *Ground-Truth Recall* designed to measure the conditional fidelity and the ground-truth fidelity of the generated samples.

First, we rasterize each generated HD map, its associated SD map used as control input during generation and its associated ground truth HD map into an RGB image. All images are then embedded into a 2048-dimensional feature space via a pretrained ResNet-50. Finally, the features of the generated HD maps are compared to the features of SD maps and ground truth HD maps via cosine similarity within batches of $B$ samples.

Specifically, we define *Control Recall* as the percentage of generated HD maps for which the associated SD map control input ranks among the top-$K$ most similar embeddings in the batch. A higher value indicates that the topology of the generated HD map aligns more faithfully with its conditioning SD map.

On the other hand, *Ground-Truth Recall* is defined as the percentage of generated HD maps for which the correct ground-truth HD map ranks among the top-$K$ most similar embeddings. A higher value indicates stronger alignment between generated and real topologies, reflecting the model's ability to reproduce realistic spatial structure under conditioning.

For both metrics, we use a batch size of $B = 64$ and a neighborhood size of $K = 3$.

# 4 Experimental Setup and Results

## 4.1 Training Setup

We build upon the *nuPlan* benchmark [3], a large-scale autonomous driving dataset providing HD maps, dynamic agents, traffic lights, and ego-vehicle states across multiple cities: Las Vegas, Pittsburgh, Singapore and Boston. To assess generalization ability and to avoid contamination of the test data, we define our custom training and test splits. The training split contains samples from Las Vegas, Pittsburgh, and Singapore, while the test split contains samples from Boston. Following SLEDGE, we condition the diffusion model on a city label to capture city-specific topology priors. For spatial control, we derive SD maps from OpenStreetMap [19] by extracting road layouts and rasterizing them into our control representation. Further details on the test dataset split and SD map rasterization are provided in the supplementary material. Our training setup builds on the multi-stage procedure introduced by SLEDGE [5] and consists of three stages in total.

**Stage 1: RVAE Latent Space.** Following SLEDGE, we first train a raster-to-vector RVAE to encode HD map elements into a compact latent representation and decode latents back to structured HD map vectors. This RVAE defines the latent space in which diffusion is performed: during Diffusion Transformer (DiT) training, ground-truth HD maps are encoded into latents, noise is added, and the diffusion model learns to denoise in this learned latent space. We keep the RVAE architecture and training protocol unchanged from [5].

**Stage 2: Unconditional DiT Baseline.** Next, we train DiT to generate HD maps *without* spatial conditioning, as in SLEDGE. Concretely, the DiT takes as input Gaussian noise in the RVAE latent space (and optionally the city label) and learns to synthesize realistic HD map latents that can be decoded by the RVAE. We train both *DiT-B* ($\sim$138M parameters) and *DiT-XL* ($\sim$487M parameters) variants to reproduce the SLEDGE baseline and to obtain a strong pretrained backbone for our controlled model. To improve robustness when the city is unknown, we randomly drop the city-label conditioning with probability 0.2 during training, so the model also learns to generate without an explicit city identifier.

**Stage 3: SD-Conditioned ControlNet on DiT.** Finally, we introduce explicit spatial conditioning by integrating an SD map-conditioned ControlNet branch into the frozen pretrained DiT backbone from the previous stage. As laid out in Sec. 3.2, we add (i) a ControlNet encoder that processes the rasterized SD control map, and (ii) ControlNet blocks that inject this control signal into the DiT denoising network. During training, each HD map is paired with its corre-sponding SD map, and the model learns to use the SD input to guide generation toward the desired road layout while recovering lane-level detail. As in stage 2, we retain the city-label input with dropout.

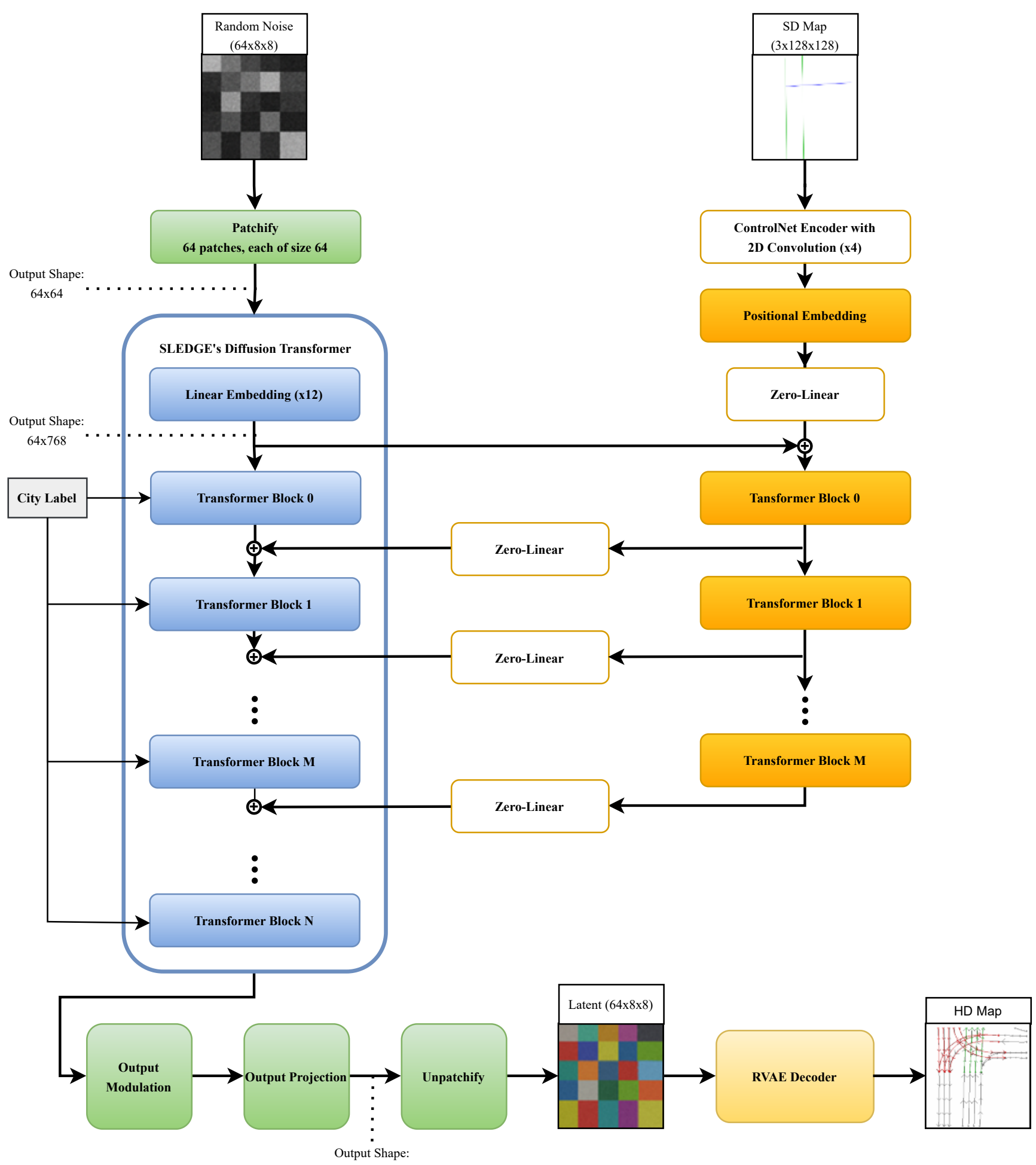

**Fig. 2:** This figure shows an overview of our pipeline for controllable HD map generation. While maintaining the entire base diffusion transformer frozen, we add a trainable ControlNet branch [27] that encodes SD maps to the SLEDGE diffusion model [5]. *Blue, green, and yellow blocks denote the original SLEDGE backbone, while orange blocks highlight our added ControlNet pathway and its residual injections.* Furthermore, city labels are input via the diffusion transformer's Adaptive Normalization layers [25]. This enables the pipeline to incorporate both spatial control and semantic conditioning for HD map synthesis. Design of control injection follows control mechanisms akin to PixArt [4] and StableAudio-Control [12].

### 4.2 Results and Discussion

To evaluate our model, we generate one HD map per sample from our Boston-only test split under two settings. We first perform unconditional generation, following the original SLEDGE [5] procedure. For each sample, Gaussian noise is drawn in the RVAE latent space and denoised by the DiT backbone without additional conditioning signals and without the city label. The resulting latent representation is then decoded by the RVAE decoder to obtain the final HD map. This setup serves as our baseline and is evaluated using reproduced DiT-B and DiT-XL checkpoints trained with our own pipeline and consistent with the original SLEDGE experiments.

We then perform conditional generation using our ControlMap pipeline. The process remains identical except that the corresponding SD map is retrieved and provided as conditioning input during denoising, allowing us to isolate the effect of SD-map guidance while keeping all other components unchanged.

As shown in Fig. 1 and Fig. 3, the model is able to generate meaningful and well-structured road layouts that fit the given spatial conditioning, although samples from Boston were held-out during training. This demonstrates that our model can generalize to urban contexts that have not been seen before while maintaining guidance alignment. Samples produced by the unguided DiT-B model, on the other hand, show arbitrary patterns and structures that do not follow spatial guidance, as shown in Fig. 3.

**Evaluating Topological Realism.** Following SLEDGE, we use the Fréchet Distance [6,17] over four topological metrics (Connectivity, Density, Reach, and Convenience). These metrics judge the quality of local structural detail and the navigability of the generated lane structure. A lower value indicates closer alignment between the distributions of generated and reference maps.

As shown in Tab. 1, our model achieves superior performance in Reach and Convenience, while falling behind the original SLEDGE model in Connectivity and Density. Fig. 4 illustrates a key reason why the unguided SLEDGE models achieve more competitive Fréchet Distances on *Connectivity* and *Density.* Without spatial conditioning, the model has considerable freedom to introduce small stubs, short spurs, and minor branch fragments. These micro-structures are frequent in the ground-truth distribution, and the model learns to reproduce them statistically. As a result, unguided generation inflates the local count of key points, which are the start or end points of lanes or lane connectors. This reduces the measured distributional distance, despite producing less realistic topology overall.

In contrast, the ControlNet-guided model follows the coarse SD map skeleton. Extracted SD maps do not contain subtle slip roads, internal connector arcs, and small branching artifacts. Consequently, our ControlNet-guided outputs preserve cleaner, smoother roads and reduce spurious micro-branches. This leads to more realistic maps, but slightly higher Fréchet Distance on metrics that reward local key point proliferation (Connectivity/Density). On global navigability metrics (*Reach* and *Convenience*), ControlNet consistently improves performance. Paths connect correctly and the global flow of traffic becomes structurally coherent, properties that derive naturally from the SD skeleton.

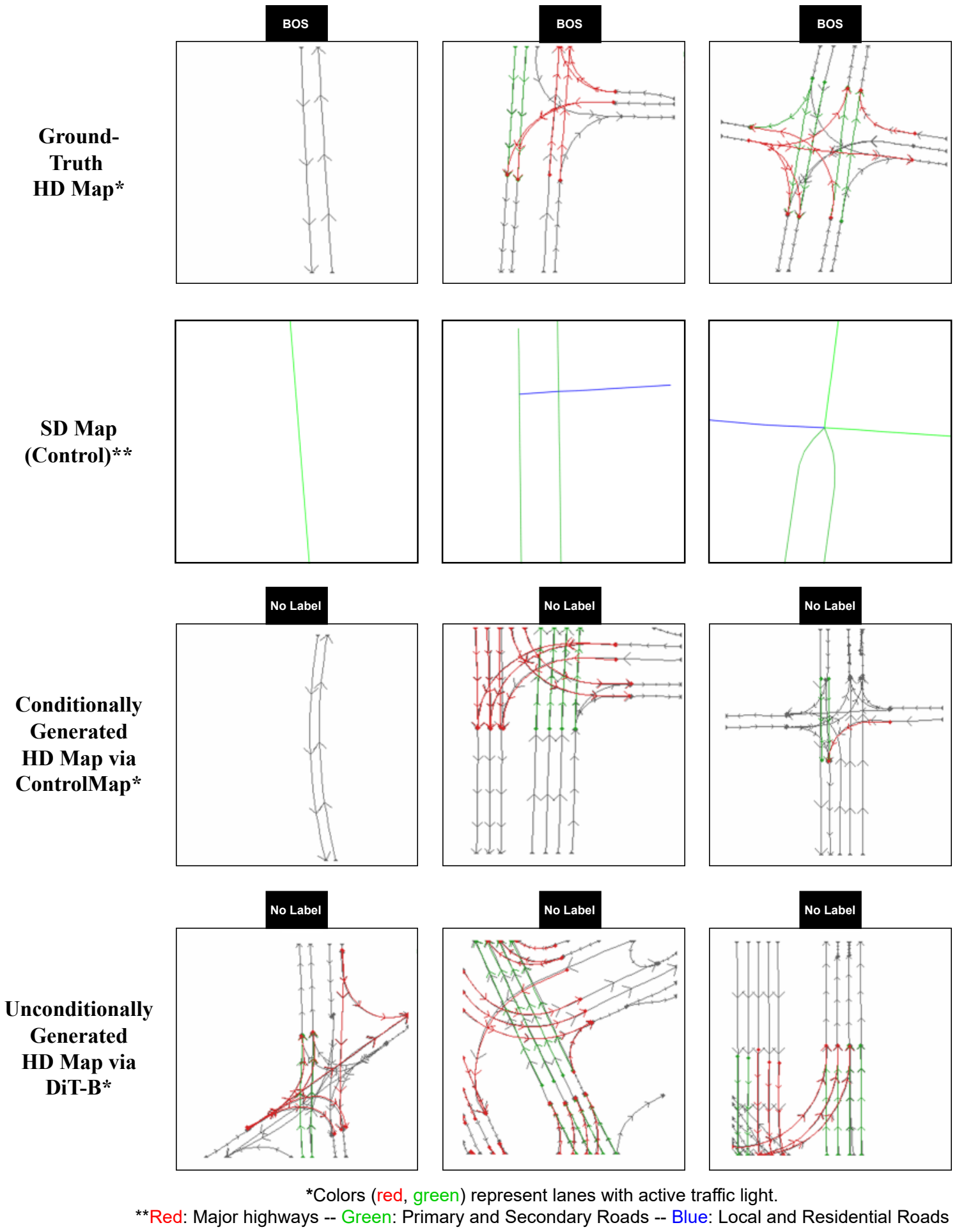

**Fig. 3:** We condition on SD maps (second row) that are cheap to obtain (*e.g.*, from OpenStreetMap [19]) to generate diverse HD maps with user-specified structure. Our generations (third row) follow the control layout (second row) and recover realistic HD detail consistent with test ground truth (first row). While the DiT-B backbone produces visually plausible HD maps (fourth row), the generation is not controllable. *BOS* denotes ground-truth maps coming from Boston; *No Label* indicates that the model took no city label as input during generation.

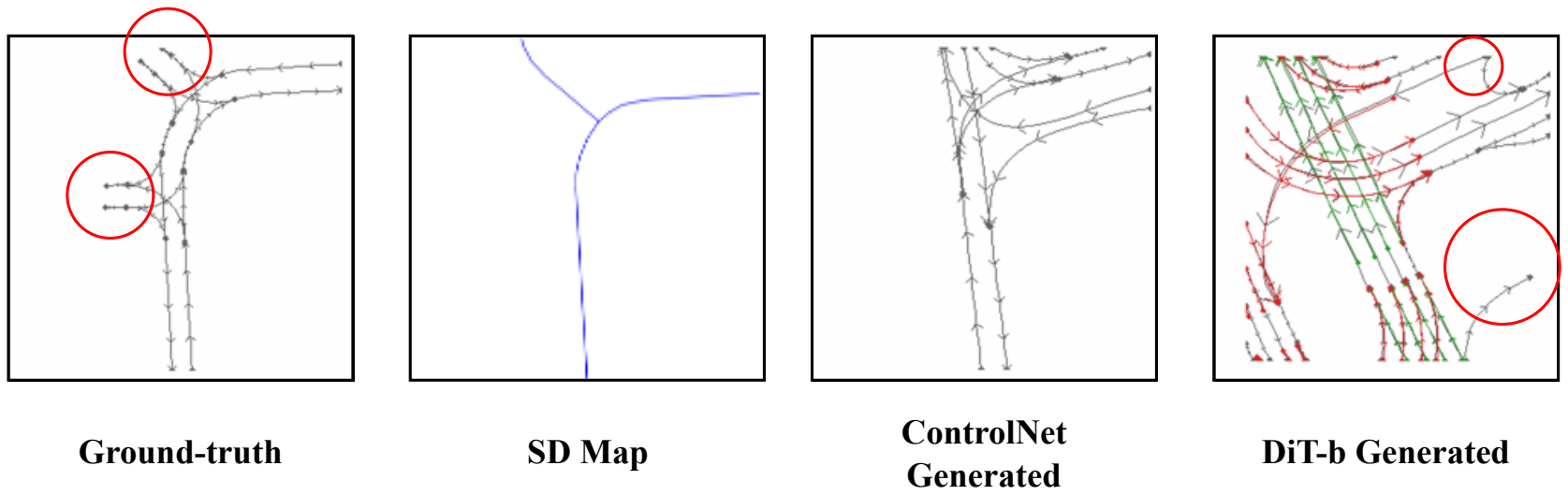

**Fig. 4:** This figure shows an illustration of SD map limitations that impact Fréchet topology metrics. ControlNet must follow the coarse structure of the SD map, which may omit minor connectors (cf. red circles), affecting topological metrics like connectivity and density. On the other hand, unguided models have full flexibility but lack global structural cues, negatively affecting metrics like convenience and reach.

**Table 1:** Fréchet Distance on topological metrics capturing topological realism

| Model | Connectivity ↓ | Density ↓ | Reach ↓ | Convenience ↓ |
|---|---|---|---|---|
| SLEDGE DiT-B | **0.0242** | 2.4127 | 0.5570 | 2.7609 |
| SLEDGE DiT-XL-Model | 0.0530 | **2.0578** | 0.6188 | 2.7836 |
| ControlMap DiT-B (ours) | 0.5453 | 2.6184 | **0.1733** | **0.7031** |

These results can partially be explained by the inherent differences between SD and HD map representations. SD maps originate from graph-structured data containing rich semantic and topological information, but converting them into rasterized control images inevitably removes part of this structure, including explicit lane connectivity and routing relationships. Furthermore, SD maps often omit fine-grained details such as minor lane connectors, lane boundaries, and traffic-control elements that are present in HD maps. As illustrated in Fig. 4, the ground-truth HD map contains some small connector roads and branching lane structures that are absent in the corresponding SD map. Consequently, the generated output follows the global road geometry provided by the SD conditioning signal, while failing to reconstruct certain local topological details that were never present in the input representation. As a result, some discrepancies between generated and ground-truth HD maps are expected, particularly in structurally ambiguous regions.

Overall, these results reflect a principled design trade-off: SD map conditioning prioritizes large-scale realism (Reach/Convenience) at the expense of some local branching complexity (Connectivity/Density), which is not explicitly encoded in the control signal.

**Feature-Space Fidelity and Diversity.** In addition to the distribution-level metrics discussed above, SLEDGE also measures two metrics called Improved Precision and Recall [15]. Improved Precision reflects the plausibility of generated samples, while Improved Recall reflects coverage of realistic topology diversity.

As shown in Tab. 2, our model with SD Map conditioning achieves consistently higher values compared to both variants of the unguided SLEDGE model. A potential explanation is that the structural prior from the SD map helps the model to generate HD maps that more closely aligns with the ground-truth distribution. It reduces mode collapse and improves coverage of rare but realistic road layouts that come from the spatial conditioning.

Despite having less than half as many parameters (roughly 230 million vs. 487 million), the ControlNet-enhanced DiT-B model performs better than even the larger DiT-XL SLEDGE model. This demonstrates how the generative quality and structural accuracy of synthetic HD maps are greatly impacted by explicit spatial conditioning through ControlNet, allowing a smaller model to perform better than a much larger unconditioned counterpart.

**Conditional Fidelity using Recall Metrics.** To complement the analysis, we also assess conditional fidelity using *Control Recall* and *Ground-Truth Recall*, our novel metrics introduced in Sec. 3.3. These metrics quantify how closely the generated HD maps match the control input and the ground-truth maps in a ResNet-50 feature space. And as shown in Tab. 2, ControlNet outperforms unguided baselines on both metrics, reflecting stronger alignment with both the conditioning SD map and the ground truth topology reference.

Furthermore, under the null hypothesis that matches between SD map/GT HD map and generated HD map occur independently with probability $K/B$ (i.e., random top-$K$ assignment within each batch of size $B$), the number of matches $M$ in a test set of $N$ samples follows a Binomial distribution: $Binom(M|N, K/B)$. Hence, the $p$-value for an observed Recall value of $r = M/N$ is given by:

$$P(R \geq r|H_0) = 1 - F_{Binom}(M-1|N, K/B),$$

where $F_{Binom}$ is the cumulative distribution function (one-sided, upper tail). With the parameters of our test setup $K = 3$, $B = 64$ and $N = 3904$, all $p$-values for the two unguided baselines are above 0.2, which is not statistically significant. On the other hand, for ControlMap, the $p$-value for Control Recall is $3.1 \times 10^{-6}$, and the $p$-value for GT Recall is below machine precision.

### 4.3 Ablation Experiments

**Effect of Classifier-Free Guidance (CFG).** CFG is a sampling technique that amplifies the influence of the conditioning signal at inference time by interpolating between unconditional and conditional model predictions [11]:

$$\epsilon_{\text{guided}} = \epsilon_{\text{uncond}} + w\left(\epsilon_{\text{cond}} - \epsilon_{\text{uncond}}\right) \tag{1}$$

where $\epsilon_{\text{uncond}}$ and $\epsilon_{\text{cond}}$ are unconditional and conditional noise predictions, respectively, and $w$ is the guidance scale. To study how stronger conditioning affects spatial adherence and map diversity, we conduct an experiment by generating a series of HD maps with fixed SD map conditioning but using CFG with $w$ set to $\{0.0, 0.3, 0.5, 0.7, 1.0\}$.

**Table 2:** This table summarizes the evaluation results across the metrics of realism, coverage, and control fidelity. Improved Precision/Recall [15] evaluate generative realism and diversity, while Control and Ground-Truth (GT) Recall assess adherence to SD conditioning and ground-truth topology. Number in brackets represent $p$-values, where $< \varepsilon$ indicates that the value is below machine precision.

| | **Realism & Coverage** | | **Control fidelity** | |
|---|---|---|---|---|
| **Model** | **Prec. ↑** | **Rec. ↑** | **Ctrl Rec. ↑** | **GT Rec. ↑** |
| SLEDGE DiT-B | 0.2823 | 0.3471 | 0.0476 (0.42) | 0.0489 (0.28) |
| SLEDGE DiT-XL | 0.2843 | 0.3186 | 0.0425 (0.91) | 0.0497 (0.21) |
| ControlMap DiT-B (ours) | **0.4390** | **0.4705** | **0.0630** **($3 \times 10^{-6}$)** | **0.2241** **($< \varepsilon$)** |

Fig. 5 illustrates how varying the guidance scale $w$ affects controllability. At $w = 0.0$, the model ignores the SD map, producing spatially unguided maps. As $w$ increases to 0.3–0.5, generated maps align progressively better with the control input. Roads become more continuous, and intersections align consistently with the SD structure. Higher values ($w \geq 0.7$) yield near-deterministic adherence to the control map, closely matching the ground-truth topology. Overall, increasing CFG strengthens semantic alignment with the SD map. This provides a simple and effective knob for trading off control fidelity against generative diversity.

**City Label Perturbation.** To analyze semantic control, we conduct an additional experiment by fixing the SD map input but exchanging the city label for a number of samples from Singapore and Las Vegas. These two cities were chosen because they exhibit opposite traffic directions (left-hand vs. right-hand driving), providing a strong probe of what the model learns from the city embedding during training, such as lane orientation and road flow patterns. As shown in Fig. 6, the city-label swaps successfully induce city-specific changes in the local detail of the generated HD maps (*e.g.*, traffic direction), supporting cross-city generalization by combining city-level priors from the label with spatial layout control from the SD map.

## 5 Conclusion

We present a generative framework for the controllable generation of HD maps that serve as a starting point for traffic scenario simulation. The architecture integrates ControlNet into an unconditioned latent diffusion model developed by [5], opening the possibility to generate maps that not only look realistic but also exhibit high conditional fidelity and adherence to readily available SD maps.

Results show that ControlNet-guided samples closely follow the provided SD map controls. In addition, we present ablation studies which demonstrate the ability to change the strength of the guidance, revealing a controllable trade-off between high conditioning strength for higher fidelity and low conditioning strength for higher diversity. Cross-city generalization tests further show that the model adapts road geometry and lane structure to different urban contexts.

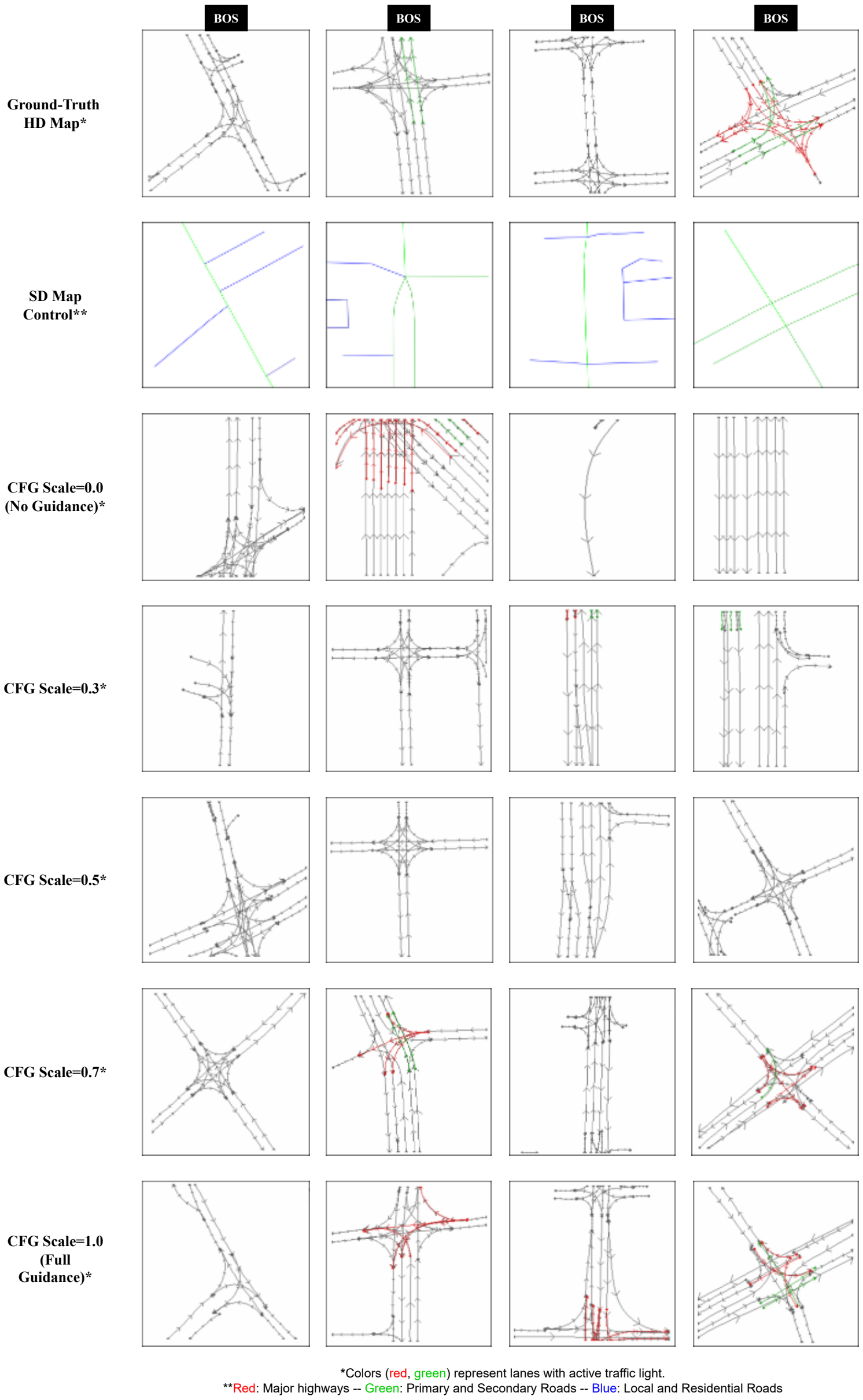

**Fig. 5:** This figure shows the effect of classifier-free guidance (CFG) on controllability. Increasing CFG improves topological alignment with the SD map, confirming controllable generation. *BOS* stands for ground-truth maps coming from Boston.

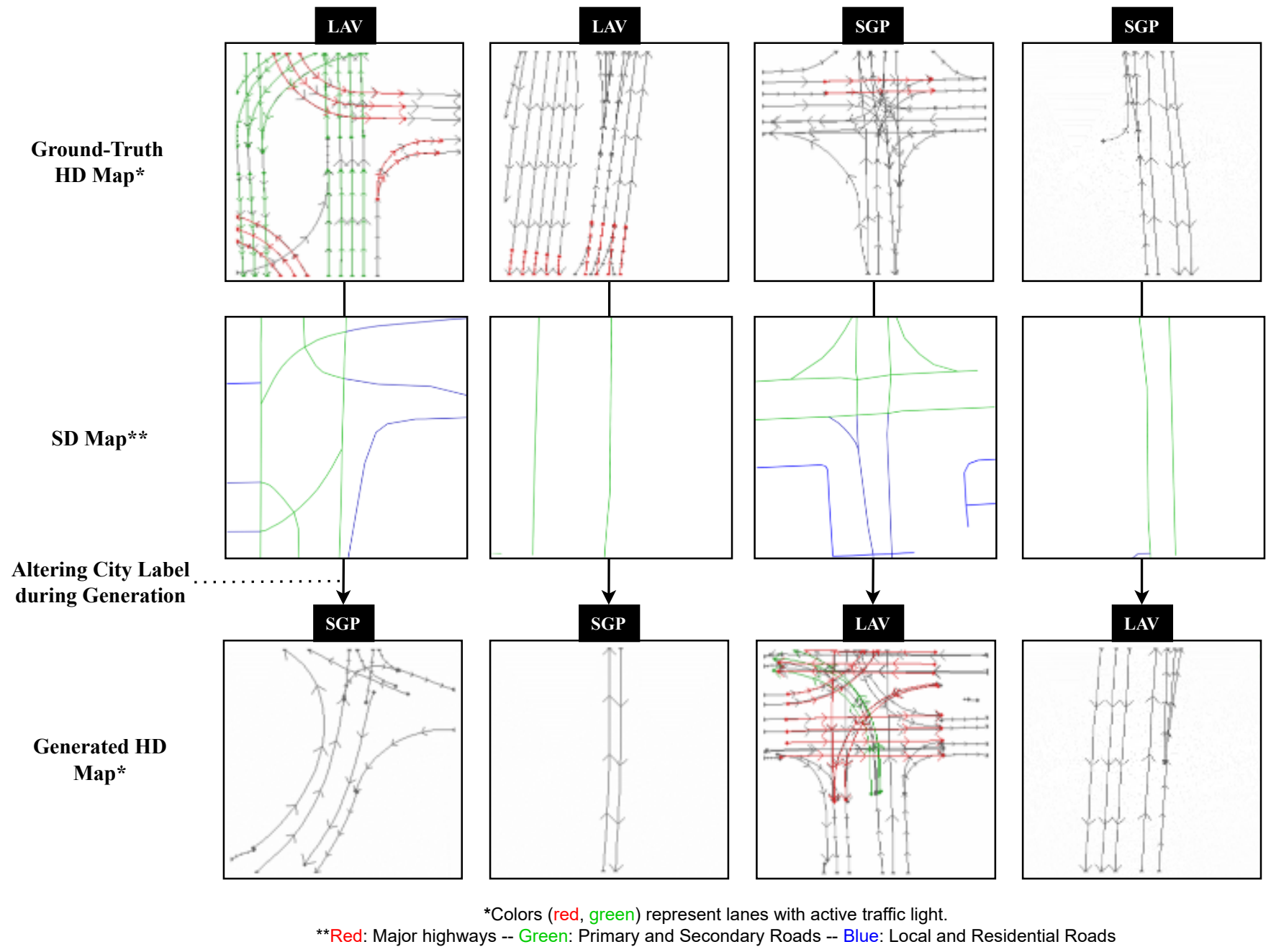

**Fig. 6:** This figure shows city-label perturbation ablation results. When swapping city labels, the model adjusts lane orientation and structure to reflect the injected city's traffic directions (left-handed vs right-handed traffic), demonstrating controllable, semantically aware generation. LAV stands for Las Vegas while SGP stands for Singapore.

This work is a key enabler for simulation based verification of automated driving systems, since it provides a cheap option to obtain realistic synthetic HD maps which adhere to a specific road topology of choice, effectively removing the dependence of simulation environments on costly real-world HD maps.

## 6 Limitations and Future Work

While SD maps enable effective spatial control, several limitations remain:

1. **Cross-source misalignment:** OSM SD maps and nuPlan HD maps may disagree (*e.g.*, missing road segments). This can be mitigated through improved preprocessing, including geometric registration, graph-based matching, and intersection-aware alignment, as well as filtering ambiguous regions using consistency thresholds. Integrating complementary modalities, such as aerial or satellite imagery, is another promising direction.

2. **Interpretation of Recall Metrics:** Control Recall and Ground-Truth Recall are proxy measures that depend on the rasterization and embedding pipeline rather than directly capturing topological similarity. Their values are sensitive to the feature extractor, batch composition, and top-$K$ selection. Future work will explore topology-aware and simulator-driven evaluation protocols for more direct assessment of structural validity.
3. **Downstream Simulation Considerations:** Further validation is required to assess the usability of generated HD maps in autonomous driving pipelines. Future work must ensure generated maps satisfy planner requirements, including graph validity, lane connectivity, routing, and agent initialization. Integrating ControlMap into closed-loop simulation environments will enable evaluation of planner robustness, traffic realism, and controllable rare-scenario generation.